\documentclass{article}
\usepackage{perpage} 
\usepackage{cite}
\MakePerPage{footnote} 

\PassOptionsToPackage{numbers, compress}{natbib}

\usepackage[final]{neurips_2021}

\usepackage[utf8]{inputenc} 
\usepackage[T1]{fontenc}    
\usepackage{url}            
\usepackage{booktabs}       
\usepackage{amsfonts}       
\usepackage{nicefrac}       
\usepackage{microtype}      
\usepackage[dvipsnames, svgnames, x11names]{xcolor}         
\usepackage{adjustbox}
\usepackage{mmstyle}
\usepackage{multirow}
\usepackage{tabularx, makecell}
\usepackage{subfig}
\usepackage{rotating}
\usepackage{comment}
\usepackage{tablefootnote}
\usepackage{CJKutf8}
\usepackage{ulem}
\usepackage{float}
\usepackage{enumitem} 

\usepackage{xcolor}
\usepackage{hyperref}
\hypersetup{colorlinks=true,
    breaklinks=true,
    urlcolor=magenta,
    linkcolor=red,
    bookmarksopen=false,
    filecolor=black,
    citecolor=green,
    linkbordercolor=red,
}

\usepackage{threeparttable}

\newcommand{\blue}[1]{\textcolor{blue}{#1}}
\newcommand{\gray}[1]{\textcolor{gray}{#1}}

\newcommand{\model}{InternVideo}

\def\etc{etc\@\xspace}

\title{InternVideo: General Video Foundation Models \\via Generative and Discriminative Learning}

\usepackage{hyperref}
\hypersetup{
    colorlinks,
    linkcolor={red!90!black},
    citecolor={blue!90!black},
}
\author{
Yi Wang$^{1*}$, 
Kunchang Li$^{5,1*}$, 
Yizhuo Li$^{3,1*}$, 
Yinan He$^{1*}$, 
Bingkun Huang$^{2,1*}$, 
Zhiyu Zhao$^{2,1*}$,
Hongjie Zhang$^{1*}$
\vspace{+1mm} \\ 
\textbf{Jilan Xu$^{4,1}$, 
Yi Liu$^{5,1}$, 
Zun Wang$^{8,1}$, 
Sen Xing$^{6,1}$, 
Guo Chen$^{2,1}$, 
Junting Pan$^{7,1}$, 
Jiashuo Yu$^{1}$} 
\vspace{+1mm}\\ 
\textbf{Yali Wang$^{5,1*}$, 
Limin Wang$^{2,1*}$, 
Yu Qiao$^{1\dagger}$}
\vspace{+3mm}\\ 
$^1$Shanghai AI Laboratory, 
$^2$Nanjing University, 
$^3$The University of Hong Kong, 
$^4$Fudan University\\
$^5$Shenzhen Institute of Advanced Technology, Chinese Academy of Sciences\\
$^6$Tsinghua University,
$^7$The Chinese University of Hong Kong, 
$^8$The Australia National University\\\\
\href{https://github.com/OpenGVLab/InternVideo}{https://github.com/OpenGVLab/InternVideo}\vspace{-5mm}
}

\begin{document}
\bibliographystyle{unsrt} 

\maketitle

\begin{figure*}[htb]
\begin{center}
  \includegraphics[width=0.89\linewidth]{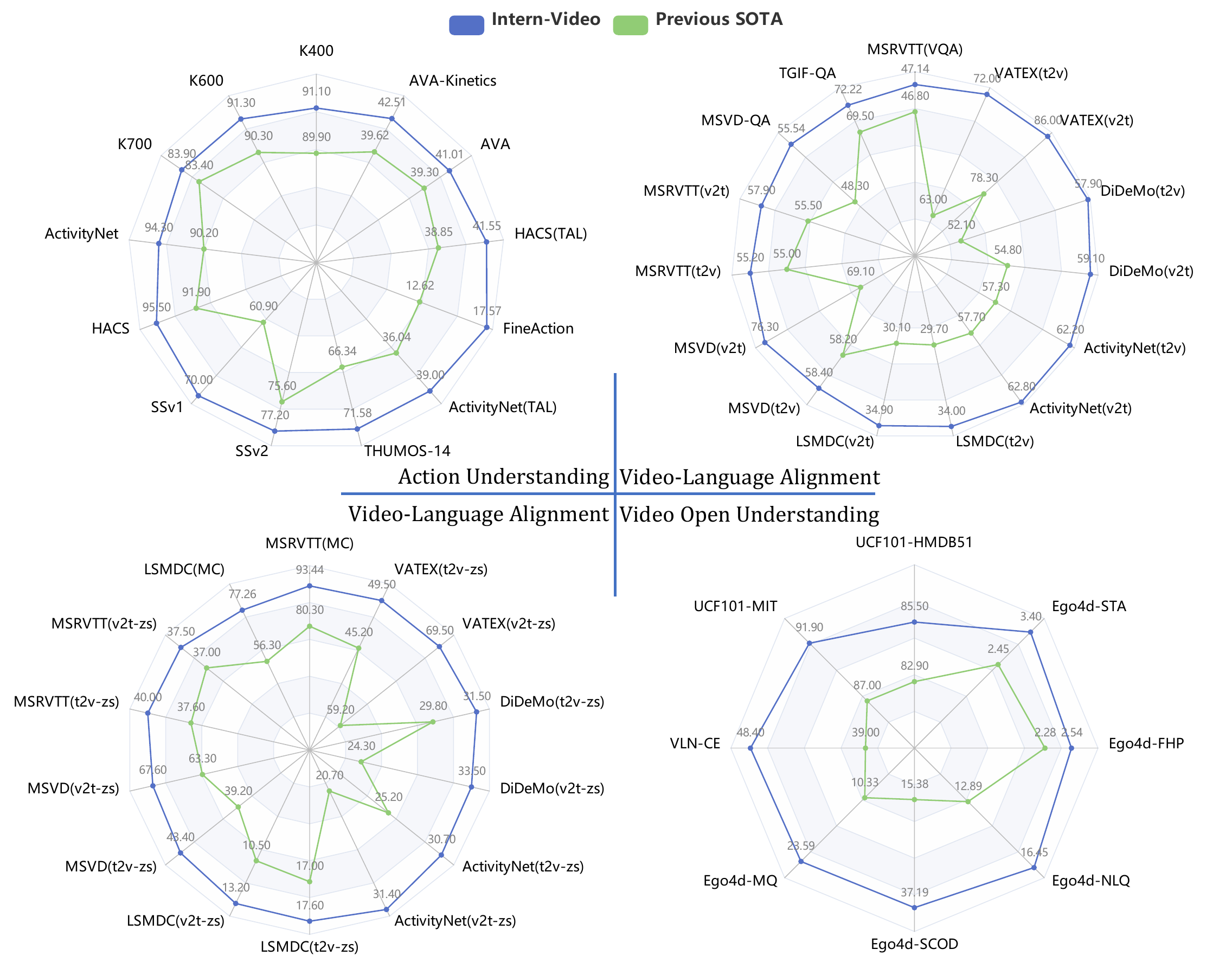}
\end{center}
  \caption{\model~delivers the best performance on extensive video-related tasks, compared with the state-of-the-art methods (including specialized \cite{xia2022nsnet,stergiou2021learn,wang2021self,zhang2022actionformer,clip4clip} and foundation models \cite{maskfeat,coca,mtv,merlot}). Comparison details are given in Section \ref{sec_tasks}. v2t and t2v denote video-to-text and text-to-video retrieval respectively. STA, FHP, NLQ, SCOD, and MQ are short for Short-term Object Interaction Anticipation, Future Hand Prediction, Natural Language Queries, State Change Object Detection, and Moment Queries, respectively.
  }
\label{fig:radar}
\end{figure*}

{
  \renewcommand{\thefootnote}%
    {\fnsymbol{footnote}}
  \footnotetext[0]{* equal contribution. $\dagger$ corresponding author (\href{mailto:qiaoyu@pjlab.org.cn}{qiaoyu@pjlab.org.cn}).} }

\begin{abstract}
The foundation models have recently shown excellent performance on a variety of downstream tasks in computer vision. However, most existing vision foundation models simply focus on image-level pretraining and adpation, which are limited for dynamic and complex video-level understanding tasks. To fill the gap, we present general video foundation models, \textit{InternVideo}, by taking advantage of both generative and discriminative self-supervised video learning. Specifically, \model~efficiently explores masked video modeling and video-language contrastive learning as the pretraining objectives, and selectively coordinates video representations of these two complementary frameworks in a learnable manner to boost various video applications. Without bells and whistles, \model~achieves state-of-the-art performance on 39 video datasets from extensive tasks including video action recognition/detection,
video-language alignment, and open-world video applications. Especially,
our methods can obtain 91.1\% and 77.2\% top-1 accuracy on the challenging Kinetics-400 and Something-Something V2 benchmarks, respectively.
All of these results effectively show the generality of our \model~for video understanding. The code will be released at \href{https://github.com/OpenGVLab/InternVideo}{https://github.com/OpenGVLab/InternVideo}.

\end{abstract}


\section{Introduction}
\label{sec:introduction}

Foundation models have been gaining increasing attention in the research community \cite{bommasani2021opportunities,akbari2021vatt,shao2021intern}, since they give a practical paradigm for scaling to numerous perception tasks with surprisingly good results. Through simple adaption or zero/few-shot learning, foundation models greatly reduce downstream design and training costs using generic representations learned from web-scale data with a strong backbone of high capacity. It is expected that developing foundation models can cultivate cognition from perception, obtaining general vision capability.

Though a line of vision foundation models is proposed \cite{CLIP,align,florence,unicl,simvlm,ofa,coca,beit3,beit,barham2022pathways}, video understanding and the corresponding tasks are less explored compared with image ones, mainly used for validating that the visual features from these models are also beneficial to spatiotemporal representations. We conjecture this relatively scarce focus from the academic community is caused by 1) a high computing burden from video processing, and 2) quite a few current video benchmarks can be handled by exploiting appearance features from image backbones with accordingly temporal modeling. Specifically, for efficiency, the additional time dimension in video processing makes at least one order of magnitude higher complexity than image processing when their spatial resolutions are close and the temporal sampling ratio is usually 16. 
For some current video datasets, image features alone or with lateral temporal modules are sufficient to give decent results, especially with the rise of the multimodal model CLIP \cite{CLIP}. Its various temporal variants yield competitive or state-of-the-art performance in several core tasks \cite{clip4clip,ma2022x}. 
Regarding this, a simultaneous spatiotemporal learner does not seem like a sweet spot between research \& development cost and payback.

Moreover, the transferability of current vision foundation models is somewhat narrow considering the wide spectrum of video applications. These models~\cite{videomae,mtv,maskfeat,all_in_one} either concentrate on action understanding tasks (\eg action recognition, spatiotemporal action localization, \etc) or video-language alignment ones (\eg video retrieval, video question answering, \etc). We suppose this results from their learning schemes, as well as the lack of a comprehensive benchmark for measuring video understanding capabilities. Thus, these works~\cite{videomae,mtv,maskfeat,all_in_one} focalize a few specific tasks to demonstrate their spatiotemporal perceptions.
The community desires a general foundation model that enables a broader application domain.

In this paper, we advance video foundation model research with a cost-effective and versatile model \model. 
To establish a feasible and effective spatiotemporal representation, we study both popular video masked modeling \cite{mae,videomae} and multimodal contrastive learning \cite{CLIP,cpd}. Note that video masking modeling specializes in action understanding, and it is still worth exploring regarding its limited model scale caused by the current decoder. For multimodal contrastive learning, it embeds rich semantics into video representation while ignoring concrete spatiotemporal modeling. To address these challenges, we make these two self-supervised approaches learn at scale efficiently in modular designs.
To significantly broaden the generalization of the current video foundation models, we propose a unified representation learning with both two self-supervised training manners. To validate such a generalized representation, we propose a systematic video understanding benchmark. It involves evaluations of action understanding, video-language alignment, and open-world video applications, which we believe are three core abilities of generic video perception. Instead of introducing new data or annotations to this system, we initially choose ten representative video tasks with 39 public datasets, and categorize them into those three types. 
To our best knowledge, \model~is the first video foundation model which demonstrates promising transferability with state-of-the-art performance in all those three different types of video tasks.

In \model, we design a unified video representation (UVR) learning paradigm. It explores both masked video modeling with autoencoders (MAE) and multimodal contrastive learning for two types of representations, strengthens them by supervised action classification, and generates a more general representation based on the cross-representation learning between them.
UVR not only empirically shows video representation outperforms image one with temporal capturing significantly on core video tasks, but also is training-efficient. Its MAE exploits high redundancies in videos and trains with only a few visible tokens. Meanwhile, multimodal learning in \model~extends existing image-pretrained backbones for video contrastive training. After supervised training these two video encoders, we craft cross-model attention to conduct feature alignments between these two almost frozen encoders.

More than a unified video representation learning paradigm, we also make practices and guidelines for training large-scale video foundation models in a tractable and efficient manner. Our work contains and is not limited to 1) making VideoMAE scalable and exploring its scalability in model and data scale; 2) efficient and effective multimodal architecture design and training receipt about how to leverage existing image-pretrained backbones; 3) empirically finding features from VideoMAE and multimodal models are complementary and studying how to deduce more powerful video representations by coordinating different existing models. Specifically, 

\begin{itemize}[leftmargin=*]
	
	\item For the scalability study of VideoMAE, we show that the proper diversity and scaling-up size in training videos can improve the scalability of the used video encoder. With a new pretrained dataset in a masked autoencoder training setting, ViT lifts its action recognition performance on Kinetics-400 \cite{k400} with finetuning from 81.01\% to 85.35\% from base to large, and further reaches 86.9\% with the huge setup, surpassing the performance reported in \cite{videomae} with a notable margin. The scalability of VideoMAE enables its usage in video foundation model development.
	
	\item For reusing existing foundation models for multimodal learning, we extend an image-pretrained vision transformer \cite{dosovitskiy2020image} to video representation learning. This transfer learning requires substantial structure and optimization customizations or using local and global spatiotemporal modules for multimodal pretraining. The local module disentangles spatiotemporal modeling by consecutive and independent spatial and temporal attention computation. Meanwhile, the global module computes token interaction across space and time. Experiments show that this reuse design is effective in spatiotemporal representation learning.
	
	\item More than self-supervised pretraining, we also employ supervised action recognition to further enhance video representation. Results demonstrate action recognition is a fine source task for transferring to various downstream applications.
	
	\item To coordinate foundation models, we unify masked video encoder with multimodal one by cross-representation learning, instead of training them jointly in one formulation. Regarding the optimization of MAE and multimodal learning (MML) that may contradict each other, combining them without compromising their merits remains an open question \cite{bachmann2022multimae}. More importantly, MML with contrastive learning demands huge batches for better contrastive optimization. Adding MAE to it will inevitably lead to numerous headachy implementation issues. Considering their potential training adversaries, we train MAE and MML separately. After their training converges, we then dynamically combine their representations with the proposed cross-model attention (CMA) modules. It implements cross-attention between MAE and MML mid-level features, adaptively fusing their high-level features for prediction. In the model-level representation interaction phase, we freeze backbones trained by MAE and MML separately and only let CMA be updated in supervised learning with a few epochs. Experiments show it is a computationally tractable and efficient means to exploit both MAE and MML features. 
\end{itemize}

We validate our proposed video foundation model in 10 tasks with 39 datasets (including core tasks \eg\ action recognition, spatiotemporal action localization, video question answering, video retrieval, \etc), and it outperforms all the state-of-the-art methods in each task non-trivially. We suppose these overall superior results obtained by our approach, along with observations and analysis, set up a new baseline for the video understanding community. The empirical evidence in this paper raises the confidence that video perceptive tasks and partial high-order tasks (formulated to perceptive forms) can be well-addressed by video foundation models, serving as a performance-critical method across a spectrum of applications.

In summary, we contribute to video foundation models in the following aspects:
\begin{itemize}[leftmargin=*]
	\item We explore a general video representation paradigm with both masked and contrastive modeling, and realize this design by unifying their representation by lightweight model interaction learning in supervision. We confirm features learned by generative and contrastive training are complementary to experiments and can deliver better results than either of them trained independently.
	\item We find masked video encoder can be scalable in model and data size with proper tuning. We devise pluggable local temporal and global spatiotemporal interaction modules to reuse pretrained ViT with image-text data for multimodal learning, easing the training burden and yielding better downstream performance.
	\item We make a tentative attempt in constructing a systematic video understanding benchmark. Our general video foundation models achieve state-of-the-art performance on 39 datasets with several core tasks in this benchmark, \eg, Kinetics-400 and Something-Something v2 in action recognition. We empirically find our learned video representations outperform their rivals, dominating vision-language tasks by a large margin, especially for some image-based ones. It suggests general video representations will be a central player in video tasks. We believe the openness of our proposed methods and models will provide the research community with handy tools to foundation models and their features with easy access. 
\end{itemize}

\section{Related Work}

\textbf{Image Foundation Models.}
Most of the current vision models are only suitable for specific tasks and domains,
and they require manually labeled datasets for training.
Regarding this, recent works have proposed vision foundation models. CLIP \citep{CLIP} and ALIGN \citep{align} prepare web-scale noisy image-text pairs to train dual-encoder models with contrastive learning,
leading to robust image-text representations for powerful zero-shot transfer. 
INTERN~\citep{shao2021intern} expands the self-supervised pretraining into multiple learning stages, which use a large quantity of image-text pairs as well as manually annotated images. INTERN achieves a better linear probe performance compared with CLIP, and improves data efficiency in the downstream image tasks.
Florence \citep{florence} extends them with unified contrastive learning \citep{unicl} and elaborate adaptation models, which support a wide range of vision tasks in different transfer settings. SimVLM \citep{simvlm} and OFA \citep{ofa} train encoder-decoder models with generative targets and show competitive performances on a series of multimodal tasks. Besides, CoCa \citep{coca} unifies contrastive learning as CLIP and generative learning as SimVLM. Recently, BeiT-3 \citep{beit3} introduces Multiway Transformers with unified BeiT \citep{beit} pretraining, achieving state-of-the-art transfer results on several vision and image-language tasks.

\textbf{Video Foundation Models.}
Previous image foundation models \citep{florence,coca} only show promising performance for video recognition (especially on Kinetics). As for video multimodal tasks, VIOLET \citep{violet} combines masked language and masked video modeling, All-in-one \citep{all_in_one} proposes unified video-language pretraining with a shared backbone, and LAVENDER \citep{lavender} unifies the tasks as masked language modeling. Though they perform well in multimodal benchmarks, they are trained with limited video-text data and struggle for video-only tasks, \eg\ action recognition. In contrast,
MERLOT Reserve \citep{reserve} collects 20M video-text-audio pairs to train the joint video representations with contrastive span matching, thus setting state-of-the-art video recognition and visual commonsense reasoning. Compared with image foundation models,
current video foundation models support limited video and video-language tasks,
especially for those fine-grained temporal discrimination tasks such as temporal localization.

\textbf{Self-supervised Pretraining.}
Self-supervised learning has developed rapidly recently. It focuses on designing different pretext tasks for pretraining \citep{doersch2015unsupervised,wang2015unsupervised,noroozi2016unsupervised,zhang2016colorful,feichtenhofer2022masked}, which can be mainly divided into contrastive learning and masked modeling. Contrastive learning adopts various data augmentations to generate different views of an image, 
then pulls together the positive pairs and pushes apart the negative pairs.
To maintain enough informative negative samples, previous methods depend on large memory banks or batch size \citep{wu2018unsupervised,he2020momentum,chen2020simple}.
BYOL \citep{byol} and SimSiam \citep{simsiam} eliminate the requirement of negative samples, designing elaborate techniques to avoid model collapse. As for masked modeling,
it learns rich visual representation via masked prediction based on visible context.
iGPT \citep{igpt} firstly mentions Masked Image Modeling (MIM). BeiT \citep{beit} propose visual token prediction with the pretrained tokenizer \citep{ramesh2021zero},
MaskFeat \citep{maskfeat} predicts the hand-crafted image descriptor, and MAE \citep{mae} directly reconstructs the raw pixels. For spatiotemporal representation learning,
VideoMAE \citep{videomae} and BEVT \citep{bevt} respectively extend MAE and BeiT to spatiotemporal space.

\textbf{Multimodal Pretraining.}
Starting from the development of image-text pretraining, large-scale video-text pretraining with specific downstream task finetuning has become the standard paradigm in the video-language area \citep{cpd,MiechASLSZ20,reserve,violet,videoclip,hu2022scaling,dou2022empirical,shen2021much,yao2021filip}. The seminal methods \citep{videobert,actbert} use pretrained visual and language encoders to extract the offline video and text features, while the recent methods \citep{cpd,MiechASLSZ20,clipbert,frozen,merlot,all_in_one} have demonstrated the feasibility of end-to-end training.  Besides, the popular methods often include two or three pretraining tasks,
\eg\, masked language modeling \citep{lavender},  video-text matching \citep{all_in_one}, video-text contrastive learning \citep{videoclip} and video-text masked modeling \citep{violet}.

\begin{figure*}[t]
	\centering
	\includegraphics[width=0.9\textwidth]{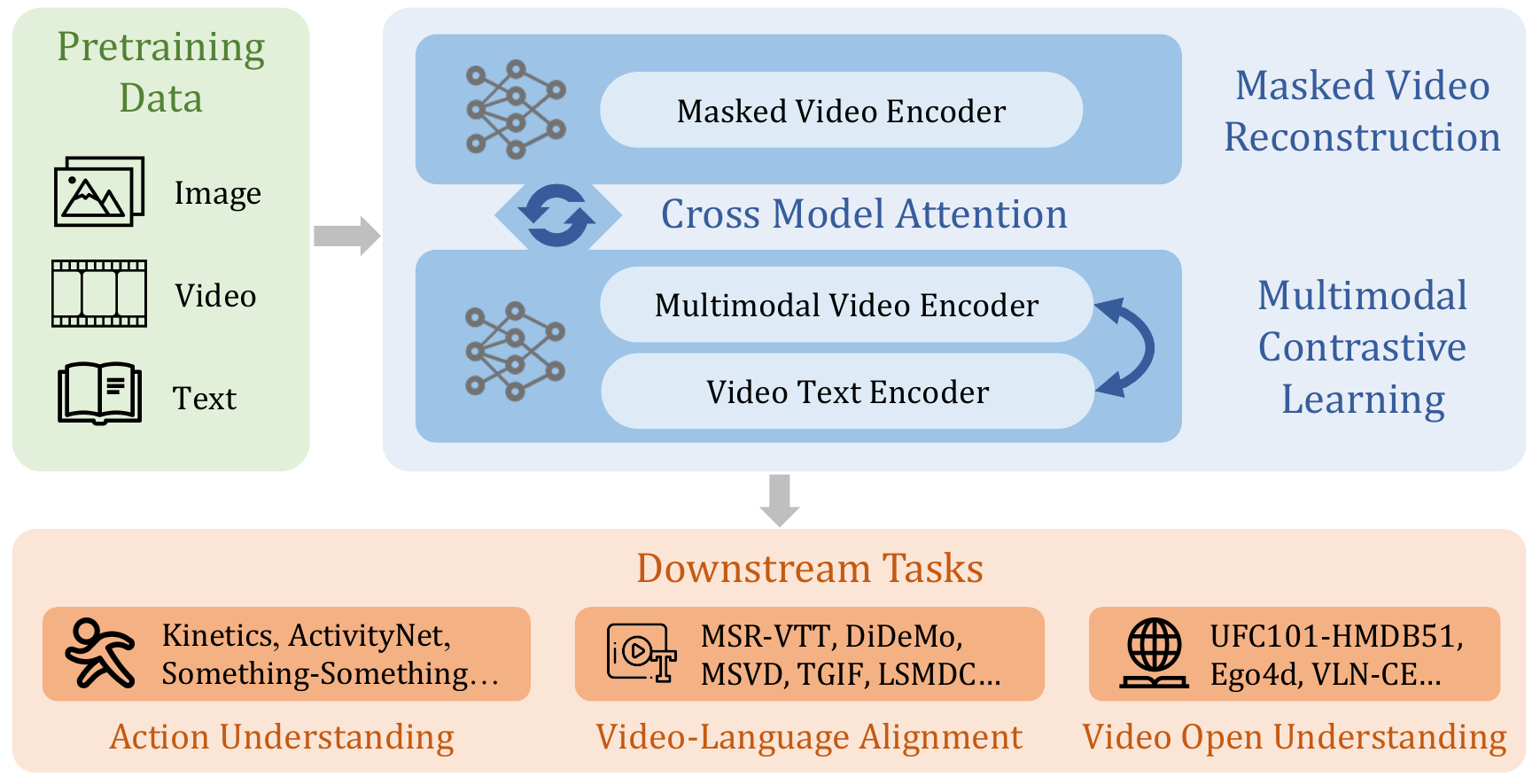}
	\vspace{-0.3cm}
	\caption{
		The overall framework of \model.
	}
	\label{frame_intern}
	\vspace{-0.6cm}
\end{figure*}
	
	\section{InternVideo}
	
	\model~is a general video foundation model along with its training and internal cooperation as given in Figure \ref{frame_intern}. In structure, \model~adopts the vision transformer (ViT) \cite{dosovitskiy2020image} and its variant UniformerV2 \cite{uniformerv2}, along with extra local spatiotemporal modeling modules for multi-level representation interaction. In learning, \model~improve its representation progressively, integrating both self-supervised (masked modeling and multimodal learning) and supervised training. Moreover, as we explore two types of self-supervised learning, we further integrate their merits. \model~dynamically derives new features from these two transformers via learnable interactions, getting the best of both worlds from generative and contrastive pertaining. Through the newly aggregated features, \model~sets new performance records on 34 benchmarks from 10 mainstream video tasks, and wins championships of five tracks in recent Ego4D competitions \cite{chen2022ego4d}.

	\subsection{Self-Supervised Video Pretraining}
	
	\model~conducts both masked and contrastive training without supervision for representation learning. According to \cite{videomae,CLIP}, video masked modeling produces features that excel at action discrimination, \eg, action recognition and temporal action localization, and video-language contrastive learning is able to understand videos with semantics from text without annotations. We employ two transformers with different structures for better leveraging these two optimization targets. The final representation is constructed by adaptively aggregating these two types of representations.
	
	\subsubsection{Video Masked Modeling}
	
	We follow most rituals from our proposed VideoMAE \cite{videomae} work to train a vanilla Vision Transformer (ViT) as a video encoder for spatiotemporal modeling, as given in Figure \ref{fig:framework} (a). VideoMAE conducts a video reconstruction task with highly masked video inputs, using an asymmetric encoder-decoder architecture. The used encoder and decoder are both ViTs. The channel number of the decoder is half of that of the encoder, with 4 blocks by default. Specifically, we divide the temporal strided downsampled video inputs into non-overlapping 3D patches and project them linearly into cube embeddings. Then we apply tube masking with notably high ratios (e.g. 90\%) to these embeddings and input them into the asymmetric encoder-decoder architecture to perform the masked video modeling pretraining. To characterize spatiotemporal interaction globally, we employ joint space-time attention \cite{arnab2021vivit,liu2022video} in ViT, making all visible tokens globally interact with each other. It is computationally tractable as only a few tokens are preserved for calculation. 

	\subsubsection{Video-Language Contrastive Learning}
	
\begin{figure*}[t]
		\centering
		\includegraphics[width=0.9\textwidth]{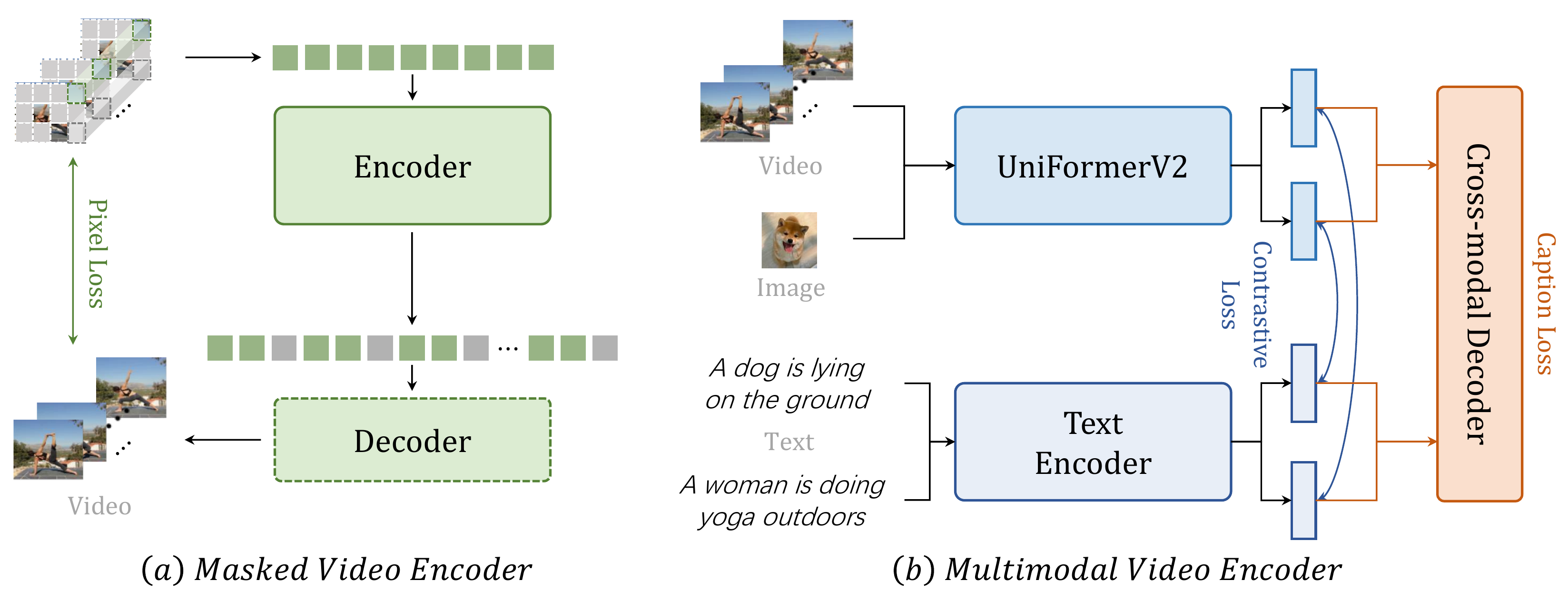}
		\vspace{-0.1cm}
		\caption{
			The overall framework of masked learning and multimodal learning in the pretrained stage.
		}
		\label{fig:framework}
	\end{figure*}

	We conduct both video/image-text contrastive learning and video captioning tasks for pretraining, as given in Figure \ref{fig:framework} (b). 
	For training efficiency,
	we build our multimodal structure based on the pretrained CLIP \cite{CLIP}. Instead of directly employing a vanilla ViT, we use our proposed UniformerV2~\citep{uniformerv2}  as the video encoder for better and more efficient temporal modeling. Moreover, we adopt an extra transformer decoder for cross-modal learning. 
	Specifically, we follow a typical align-before-fuse paradigm as given in \cite{coca,li2021align}. 
	First, video and text are separately encoded. Then a contrastive loss is utilized to align the embedding space of video and text features. In the fusing stage, we apply a caption decoder as a cross-modality fuser, which uses cross attention for a captioning pretext. This align-before-fuse paradigm not only ensures the modalities can be aligned into the same single embedding space, which is beneficial for tasks like retrieval but also gifts the model with the ability to combine different modalities and can be beneficial for tasks like question answering. The introduction of a caption decoder both extends the potential of the original CLIP and improves the robustness of multimodality features.
	
	\subsection{Supervised Video Post-Pretraining}
	
	Empirically, action recognition acts well as a meta task in video downstream applications, widely validated in \cite{lin2018bsn,lin2019bmn}. Thus we train a masked video encoder and a multimodal one with supervised action classification separately as a post-pretraining step for better performance in diverse tasks. To promote the learning capacity of these encoders, we propose a unified video benchmark Kinetics-710~(K710, described in Section \ref{k710}) for finetuning our video encoders.
	
	\textbf{Masked Video Encoder.} We finetune the masked video encoder with 32 GPUs on K710. We adjust the learning rate linearly according to the base learning rate and batch size, $\textit{lr} = \textit{base learning rate} \times \frac{\textit{batch size}}{256} $. We adopt DeepSpeed\footnote{\url{https://github.com/microsoft/DeepSpeed}} framework to save memory usage and speed up training. We set the base learning rate to $0.001$, the drop path rate to $0.2$, the head dropout rate to $0.5$, the repeated sampling\cite{Hoffer2020AugmentYB} to 2, the layer decay to $0.8$, and trained for $40$ epochs.

	\textbf{Multimodal Video Encoder.}
	We follow most of the training recipes in UniFormer \cite{li2022uniformer}. For the best result, we adopt CLIP-ViT \cite{CLIP} as the backbone by default, due to its robust representation pretrained by vision-language contrastive learning.
	We insert the global UniBlocks in the last 4 layers of ViT-B/L to perform the multi-stage fusion.
	We set the base learning rate to $ 1e -5 $, the repeated sampling to 1, the batch size to 512, and trained for 40 epochs. We adopt sparse sampling \citep{wang2016temporal} with a resolution of 224 for all the datasets.
	In post-pretraining, we use a UniformerV2~\cite{uniformerv2} as the visual encoder and initialize additional parameters in a way that the output is identical to the original CLIP model which we find to be essential for good zero-shot performance. The video captioning module is a standard 6-layer transformer decoder with $c=768$ followed by a two-layer MLP. Other setting leaves CLIP Large/14 untouched. 
	
	\begin{figure*}[t]
		\centering
		\includegraphics[width=0.9\textwidth]{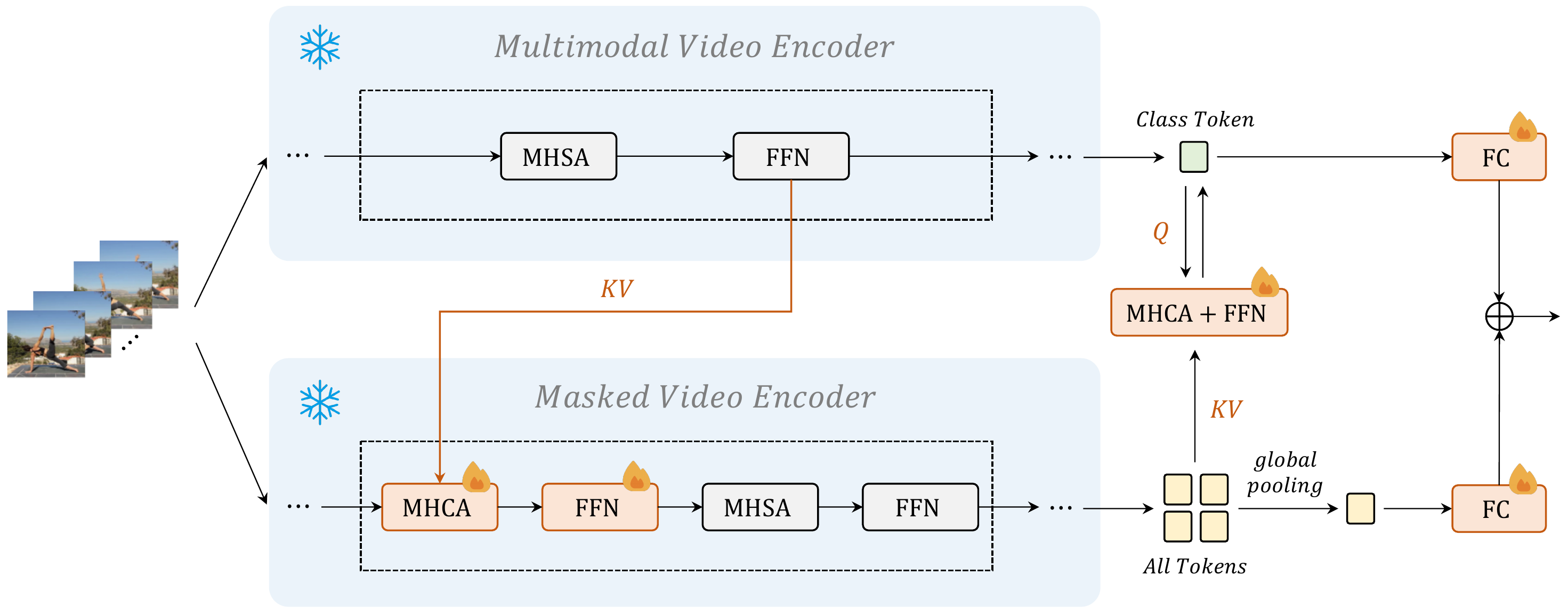}
		\vspace{-0.1cm}
		\caption{
			The illustration of the model interaction using cross-model attention.
		}
		\label{fig:ensemble}
		\vspace{-0.4cm}
	\end{figure*}
	
	\subsection{Cross-Model Interaction}
	
	To learn a unified video representation based on both video masked modeling and video-language contrastive learning, we conduct cross-representation learning with added cross-model attention modules, as shown in Figure \ref{fig:ensemble}. 
	
	Regarding optimizing both models at the same time is computing-intensive, 
	we freeze both backbones except the classification layers and the query tokens in the multimodal video encoder, only updating newly added components.
	We add some elaborate learnable modules (cross-model attention) for aligning representations learned in different approaches. Cross-model attention (CMA) is formed by standard Multi-Head Cross Attention (MHCA) along with Feed-Forward Network (FFN). It employs intermediate tokens from a multimodal video encoder as keys and values while using these from a masked video encoder as queries. The new tokens computed from CMA are treated as a gradually aligned representation with that from the multimodal video encoder. This procedure mainly transfers multimodal knowledge to CMAs in the masked video encoder. One design exception is that for the last CMA module, its keys and values are from the tokens of the masked video encoder and the query is from the class token of the multimodal video encoder. Thus, the class token is updated based on tokens from the masked encoder. It transfers single-modal knowledge to CMA in the multimodal video encoder. From this perspective, the features in all stages of the masked video encoder and the ones in the final stage of the multimodal video encoder are enhanced to coordinate with each other, in the supervision by action recognition. Finally, we utilize a learnable linear combination to dynamically fuse the two prediction scores.

\section{Experiments}

We detail our experimental configurations first (Section \ref{sec_config}), 
then we present the downstream performance of \model~on the proposed video understanding benchmark with three types of tasks (action understanding, video-language alignment, and open understanding) in Section \ref{sec_tasks}.

\subsection{Data for Pretraining} \label{sec_config}

General video foundation model pretraining requires data from various domains at scale. To achieve a data distribution with diversity, we employ 6 public datasets and our self-collected video clips as shown in Table~\ref{tab:pretraining_datasets}. 

\begin{table*}[t]
\centering
\caption{\label{tab:pretraining_datasets}Summary of datasets used in InternVideo pretraining process. A massive scale database is crucial to general vision pretraining. Our pretraining data is composed of 12 million video clips from 5 different domains.}
\setlength{\tabcolsep}{5.5 mm}{
\begin{tabular}{lccc}
\toprule
Pretraining Dataset   & \multicolumn{1}{l}{Domain}             & \multicolumn{1}{l}{Sample Clips} & \multicolumn{1}{l}{ Frames $\times$ Sample rate } \\ \midrule
Kinetics-400   \cite{k400}        & \multicolumn{1}{l}{Youtube Video}      & \multicolumn{1}{l}{240k}         & \multicolumn{1}{l}{$ 16 \times 4 $}               \\
WebVid2M   \cite{frozen}  & \multicolumn{1}{l}{Web Video}          & \multicolumn{1}{l}{250k}         &
\multicolumn{1}{l}{$ 16 \times 4 $}               \\
WebVid10M      \cite{frozen}        & \multicolumn{1}{l}{Web Video}          & \multicolumn{1}{l}{10M}         &
\multicolumn{1}{l}{$ 16 \times 4 $}               \\
HowTo100M   \cite{Miech2019HowTo100MLA}         & \multicolumn{1}{l}{Youtube Video}          & \multicolumn{1}{l}{1.2M}         &
\multicolumn{1}{l}{$ 16 \times 4 $}  \\
AVA        \cite{ava}     & \multicolumn{1}{l}{Movie}              & \multicolumn{1}{l}{21k}         & \multicolumn{1}{l}{$ 16 \times 4 $}               \\
Something-Something V2 \cite{goyal2017something} & \multicolumn{1}{l}{Shot from scripts}  & \multicolumn{1}{l}{169k}         & \multicolumn{1}{l}{$ 16 \times 2 $}               \\
Self-collected video   & \multicolumn{1}{l}{Youtube, Instagram} & \multicolumn{1}{l}{250k}         & \multicolumn{1}{l}{$ 16 \times 4 $}               \\
Kinetics-710 \cite{uniformerv2} & \multicolumn{1}{l}{Youtube Video} & \multicolumn{1}{l}{680k} & \multicolumn{1}{l}{$16 \times 4$}\\
\bottomrule
\end{tabular}
}
\end{table*}

\paragraph{Kinetics-710.} \label{k710}
We adopt a new customized kinetics action dataset Kinetics-710 \cite{uniformerv2} for supervised training, both separate and joint ones. It has 650K videos with 710 unique action labels. It combines all the unique training data from Kinetics 400/600/700~\cite{k400,k600,k700}. To avoid the training leak, some training data existing in the testing set from Kinetics of a specific version are abandoned.

\paragraph{UnlabeledHybrid.} \label{hybrid}
The UnlabeledHybrid dataset is used for masked video pretraining, which is consist of Kinetics-710~\cite{uniformerv2}, Something-Something V2~\cite{goyal2017something}, AVA~\cite{ava}, WebVid2M~\cite{frozen}, and our self-collected videos. For AVA, we cut the 15-minute training videos by 300 frames and get 21k video clips. We just randomly pick 250k videos from Self-collected videos and WebVid2M respectively. More details can be seen in Table.~\ref{tab:pretraining_datasets}.

\subsection{Implementations}

\subsubsection{Multimodal Training}

With the initialization from CLIP, we post-pretrain our multi-modal model with WebVid2M, WebVid10M, and HowTo100M. Since the training corpus of video-text datasets is not as rich as CLIP-400M~\cite{CLIP}, we co-train the video model with image-text datasets, a subset of LAION-400M~\cite{laion400m} containing 100M image-text pairs. We alternate images and videos for each iteration. The batch size of video-text is 14,336, and the batch size of image-text is 86,016. We train for 400k steps on 128 NVIDIA A100 GPUs in 2 weeks, with a learning rate of $8\times 10^{-5}$, weight decay of 0.2, cosine annealing schedule, and 4k warm-up steps.

\subsubsection{Masked Video Training}
We train the VideoMAE-Huge for 1200 epochs on the UnlabeledHybrid dataset with 64 80G-A100 GPUs. The model adapts the cosine annealing learning rate schedule and warmup 10\% total epochs. The learning rate is set to $2.5e-4$. Only MultiScaleCrop is used for data augmentation.

\subsubsection{Model Interaction}

As shown in Figure \ref{fig:ensemble},
we freeze both backbones except the classification layers and the query tokens in the multimodal video encoder.
To maintain the original output,
we add tanh gating layers in the extra MHCA and FFN as in Flamingo \cite{alayrac2022flamingo}, 
and the parameters in dynamic weighted sum are initialized as zero.
We train the coordinated models with a batch size of 64, a learning rate of $5\times10^{5}$, a weight decay of 0.001, a dropout rate of 0.9, and an EMA rate of 0.9999.
Besides,
we use a cosine annealing schedule for 5 epochs with 1 warmup epoch.
All used data augmentations are the same as in UniFormerV2 \cite{uniformerv2}.

\subsection{Downstream Tasks} \label{sec_tasks}
We conduct extensive experiments on a spectrum of downstream tasks to evaluate \model. The employed tasks are of three categories that consider action understanding, video-language alignment, and open understanding. Since \model~contains masked video encoder specializing in spatiotemporal variation characterization and fused multi-modality video encoder, it can improve action understanding (Section \ref{sec_action}) and video-language alignment (Section \ref{sec_vlt}) tasks significantly. Its generalization brought by large-scale training data also enables its impressive zero-shot and open-set capabilities on the related tasks (Section \ref{sec_open}). Even transferred to ego-centric tasks, \model~still gives an overwhelmingly favorable performance with simple heads \cite{chen2022ego4d}. Details are given as follows.

\begin{table}[t]
\centering
\caption{Action recognition results on Kinetics \& Something-Something. We report the top-1 accuracy of the compared methods on each dataset. \model-D indicates it is formed by the model ensemble between a masked video encoder ViT-H and a CLIP-pretrained UniFormerV2-L, while \model-T indicates it is computed based on \model-D and a multimodal-pretrained UniFormerV2-L.}
\setlength{\tabcolsep}{6.3 mm}{
  \begin{tabular}{lllll}
    \toprule
     Method & \#Params & K400 & K600 & K700 \\
    \midrule
    MaskFeat-L \cite{maskfeat} & 218M &  87.0 & 88.3 & 80.4 \\
    CoCa \cite{coca} & 1B+ & 88.9 & 89.4 & 82.7 \\ 
     MTV-H ~\cite{mtv} & 1B+ & 89.9 & 90.3 & 83.4 \\
    MerlotReserve-L \cite{merlot} & 644M & - & 90.1 & - \\
    \gray{MerlotReserve-L (+Audio) \cite{merlot}} & \gray{644M} & \gray{-} & \gray{91.1} & \gray{-} \\
    \hline
     \model-D & 1.0B & 90.9 & 91.1 & 83.8 \\
     \model-T & 1.3B & \textbf{91.1}$_{(+1.2)}$ & \textbf{91.3}$_{(+1.0)}$ & \textbf{84.0}$_{(+0.6)}$ \\
    \bottomrule
  \end{tabular}
}
\label{tab:kinetics_emsemble}
\end{table}

\begin{table}[t]
\centering
\caption{Action recognition results on Something-Something \& ActivityNet \& HACS \& HMDB51. We report the top-1 accuracy of the compared methods on each dataset.}
\setlength{\tabcolsep}{5.5 mm}{
  \begin{tabular}{llllll}
    \toprule
     Method & SthSthV1 & SthSthV2 & ActivityNet & HACS & HMDB51\\
    \midrule
    Previous SOTA & 60.9 \cite{ean} & 75.6 \cite{maskfeat} & 90.2 \cite{xia2022nsnet}  & 91.9 \cite{arnab2021vivit} & 87.6 \cite{wang2021self} \\
     \model & \textbf{70.0}$_{(+9.1)}$ & \textbf{77.2}$_{(+1.6)}$ & \textbf{94.3}$_{(+4.1)}$ & \textbf{95.5}$_{(+3.6)}$ & \textbf{89.3}$_{(+1.7)}$ \\
    \bottomrule
  \end{tabular}
}
\label{tab:act}
\end{table}

\begin{table}[!h]
\centering
\caption{Temporal action localization results on THUMOS-14 \& ActivityNet-v1.3 \& HACS \& FineAction. We report the average mAP of the compared methods on each dataset.}
\setlength{\tabcolsep}{3.2 mm}{
  \begin{tabular}{llllll}
    \toprule
     Backbone & Head & THUMOS-14 & ActivityNet-v1.3 & HACS & FineAction \\
    \midrule
    I3D \cite{k400} & ActionFormer \cite{zhang2022actionformer} & 66.80 & - & - & 13.24 \\
    SlowFast \cite{feichtenhofer2019slowfast} & TCANet \cite{qing2021temporal} & - & - & 38.71 & - \\
    TSP \cite{alwassel2021tsp} & ActionFormer \cite{zhang2022actionformer} & - & 36.60 & - & - \\
    \hline
     \model & ActionFormer \cite{zhang2022actionformer} & \textbf{71.58}$_{(+4.78)}$ & \textbf{39.00}$_{(+2.40)}$ & {41.32} & \textbf{17.57}$_{(+4.33)}$ \\
     \model & TCANet \cite{qing2021temporal} & {-}  & {-}  & \textbf{41.55}$_{(+2.84)}$ & {-}  \\
    \bottomrule
  \end{tabular}
}
\label{tab:tal_emsemble}
\end{table}
\begin{table}[h]
\centering
\caption{Spatiotemporal action localization results on AVA2.2 \& AVA-Kinetics (AK). We report the mAP of the evaluated approaches on datasets.}
\setlength{\tabcolsep}{9.8 mm}{
  \begin{tabular}{lcll}
    \toprule
     Method & Head & AVA2.2 & AVA-Kinetics\\
    \midrule
     ACAR \cite{pan2021actor} (ensemble) & ACAR \cite{pan2021actor} 
     & 33.30 & 40.49\\
     RM \cite{feng2021relation} (ensemble)  & RM \cite{feng2021relation} 
     & - & 40.97 \\
     MaskFeat \cite{maskfeat} & - 
     &  38.80 & - \\

     \model~ & Linear 
     & 
 \textbf{41.01}$_{(+2.21)}$ & \textbf{42.51}$_{(+1.54)}$ \\
    \bottomrule
  \end{tabular}
}
\label{tab:stal}
\end{table}

\subsubsection{Action Understanding Tasks}
\label{sec_action}
\textbf{Action Recognition.}
Actions derive spatiotemporal patterns. \model~aims to learn the representation of suitable spatiotemporal features, and the modeling of dynamical patterns. We evaluate \model~on 8 action recognition benchmarks, including popular Kinetics and Something-Something.

We evaluate VideoMAE and UniFormerV2 in \model~on Kinetics-400~\cite{k400}, Kinetics-600~\cite{k600}, Kinetics-700~\cite{k700}, Something-in-Something-V1~\cite{goyal2017something}, Something-in-Something-V2~\cite{goyal2017something}, ActivityNet~\cite{caba2015activitynet}, HACS~\cite{zhao2019hacs}, and
HMDB51~\cite{hmdb51}. We use the top-1 accuracy as a comparison indicator. In Table \ref{tab:kinetics_emsemble} and \ref{tab:act}, \model~demonstrates exceedingly promising performance on all these action recognition benchmarks.
Our \model~significantly surpasses previous SOTA methods on almost all benchmarks and matches the SOTA result on ActivityNet. The rised accuracy brought by extra fused model (\model-D vs. \model-T) demonstrates it is necessary to explore a broad technical roadmap as different lines benefits each other in performance.

\textbf{Temporal Action Localization.}
This task (TAL) aims at localizing the start and end points of action clips from the entire untrimmed video with full observation.
We evaluate our InternVideo on four classic TAL datasets: THUMOS-14 \cite{idrees2017thumos}, ActivityNet-v1.3 \cite{caba2015activitynet}, HACS Segment  \cite{zhao2019hacs} and FineAction \cite{liu2022fineaction}.
In line with the previous temporal action localization tasks, we use mean Average Precision (mAP) for quantitative evaluations. Average Precision (AP) is calculated for each action category, to evaluate the proposals on action categories. It is computed under different tIoU thresholds.
We report the performance of the state-of-the-art TAL methods whose codes are publicly available, including ActionFormer \cite{zhang2022actionformer} method for THUMOS-14, ActivityNet-v1.3, and FineAction and TCANet \cite{qing2021temporal} method for HACS Segment.

We use ViT-H from our InternVideo as backbones for feature extraction. 
In our experiment, ViT-H models are pretrained from Hybrid datasets.
As shown in Table \ref{tab:tal_emsemble},
our \model~outperforms best than all the preview methods on these four TAL datasets.
Note that, our InternVideo achieves huge improvements in temporal action localization, especially in fine-grained TAL datasets such as THUMOS-14 and FineAction.

\textbf{Spatiotemporal Action Localization.}
This task (STAL) is to predict the frames and corresponding actions of people in video keyframes.
We evaluate \model~on two classic STAL datasets AVA2.2~\cite{ava} and AVA-Kinetics~\cite{avakinetics}. In AVA2.2~\cite{ava}, each video lasts 15 minutes, and it gives a keyframe every second. The annotation is provided for keyframes instead of all frames. Here we use a classic two-stage approach to handle this task. We apply a well-trained (on MS-COCO \cite{lin2014microsoft}) Mask-RCNN \cite{maskrcnn} to detect humans on each keyframe, and the keyframe boxes are provided in the Alphaction~\cite{alphaction} project. In the second stage, centering around the key frame, a certain number of frames are extracted and fed into our video backbone.
Similarly, in the training, we use the ground truth box for training~\cite{alphaction}, and the boxes predicted in the first stage for testing.

We used ViT-Huge in \model~for experiments. The specific results can be seen in Table \ref{tab:stal}. The classification head uses a simple linear head, achieving the SOTA performance on both datasets. Note using the ViT-H model, and training with the AVA-Kinetics dataset not only improves the overall mAP, but also significantly improves the mAP obtained by testing on AVA alone. It suggests the introduction of some Kinetics videos to AVA will improve the generalization of the model over AVA; on the other hand, observing the various distributions of the AVA dataset, we find that AVA presents a typical long-tailed distribution. The introduction of Kinetics video will alleviate this issue for better results. Due to the small number of models validated on the AVA-Kinetics dataset, only the results from the \href{https://paperswithcode.com/sota/spatio-temporal-action-localization-on-ava}{paperswithcode} website are selected in the Table \ref{tab:stal}.

\subsubsection{Video-Language Alignment Tasks}
\label{sec_vlt}
\begin{table}[t]
\centering
\caption{Results of video retrieval on MSR-VTT, MSVD, LSMDC, ActivityNet, DiDeMo, and VATEX. We report R@1 both on text-to-video (T2V) and video-to-text (V2T) retrieval tasks.}
\setlength{\tabcolsep}{2.5 mm}{
  \begin{tabular}{lcccccccccccc}
    \toprule
    \multirow{2}{*}{Method}& \multicolumn{2}{c}{MSR-VTT} & \multicolumn{2}{c}{MSVD} & \multicolumn{2}{c}{LSMDC} & \multicolumn{2}{c}{ActivityNet} & \multicolumn{2}{c}{DiDeMo} & \multicolumn{2}{c}{VATEX}  \\
     & T2V & V2T & T2V & V2T & T2V & V2T & T2V & V2T & T2V & V2T & T2V & V2T \\
    \midrule
     CLIP4Clip \cite{clip4clip} & 45.6 & 45.9 & 45.2 & 48.4 & 24.3 & 23.8 & 40.3 & 41.6 & 43.0 & 43.6 & 63.0 & 78.3 \\
     TS2Net \cite{liu2022ts2} & 49.4 & 46.6 & - & - & 23.4 & - & 41.0 & - & 41.8 & - & 59.1 & - \\
     X-CLIP \cite{ma2022x} & 49.3 & 48.9 & 50.4 & 66.8 & 26.1 & 26.9 & 46.2 & 46.4 & 47.8 & 47.8 & - & - \\
     \model~ & \textbf{55.2} & \textbf{57.9} & \textbf{58.4} & \textbf{76.3} & \textbf{34.0} & \textbf{34.9} & \textbf{62.2} & \textbf{62.8} & \textbf{57.9} & \textbf{59.1} & \textbf{71.1} & \textbf{87.2} \\
    \bottomrule
  \end{tabular}
}

\label{tab:video_retrieval}
\end{table}
\begin{table}[t]
    \centering
    \caption{{Video question answering on MSRVTT, MSVD, and TGIF. We report top-1 accuracy.}
    }
    \begin{tabular}[htop]{l l l l}
        \toprule
        Method  & MSRVTT & MSVD & TGIF \\
        \midrule
        ClipBERT~\cite{clipbert} & 37.4 & - &  60.3 \\
        All-in-one~\cite{all_in_one} & 42.9 & 46.5 & 64.2 \\
        MERLOT~\cite{merlot} & 43.1 & - & {69.5} \\
        VIOLET~\cite{violet} & {43.9} & {47.9} & 68.9 \\
        InternVideo & \textbf{47.1}$_{(+3.2)}$ & \textbf{55.5}$_{(+7.6)}$ & \textbf{72.2}$_{(+3.3)}$ \\
        \bottomrule
    \end{tabular}
    \label{tab:vqa}
    \vspace{-0.1cm}
\end{table}
\begin{table}[t]
	\centering
	\caption{Results of VLN-CE dataset.}
        \setlength{\tabcolsep}{2.4 mm}{
	\begin{tabular}{cclclll}
			\toprule 
			\multirow{2}{*}{Agent} &
			\multirow{2}{*}{Backbone}  & \multicolumn{5}{c}{val-unseen} \\
			\cmidrule(r){3-7}
			& & {NE$\downarrow$} & {PL} &
		    {SR$\uparrow$} & {NDTW$\uparrow$} & {SPL$\uparrow$} \\
			\midrule
			CWP-VLNBERT~\cite{hong2022bridging} & ResNet50 & 5.52 & 11.85 & 45.19 & 54.20 & 39.91 \\
			CWP-HEP~\cite{an20221st} & CLIP-ViT-B/16 & 5.21 & 10.29 & 50.24 & 60.39 & 45.71  \\
                CWP-HEP~\cite{an20221st} & \model~ & \textbf{4.95}$_{(-0.26)}$ & 10.44 & \textbf{52.90}$_{(+2.66)}$ & \textbf{61.55}$_{(+1.16)}$ & \textbf{47.70}$_{(+0.99)}$ \\
			\bottomrule
	\end{tabular}
        }
	\label{tab:vlnce}
\end{table}
\begin{table}[!th]
\centering
\caption{Results of zero-shot video retrieval on MSR-VTT, MSVD, LSMDC, ActivityNet, DiDeMo, and VATEX. We report R@1 both on text-to-video (T2V) and video-to-text (V2T) retrieval tasks.}
\setlength{\tabcolsep}{2.7 mm}{
  \begin{tabular}{lcccccccccccc}
    \toprule
    \multirow{2}{*}{Methods}& \multicolumn{2}{c}{MSR-VTT} & \multicolumn{2}{c}{MSVD} & \multicolumn{2}{c}{LSMDC} & \multicolumn{2}{c}{ActivityNet} & \multicolumn{2}{c}{DiDeMo} & \multicolumn{2}{c}{VATEX}  \\
     & T2V & V2T & T2V & V2T & T2V & V2T & T2V & V2T & T2V & V2T & T2V & V2T \\
    \midrule
     CLIP \cite{CLIP} & 35.0 & 32.3 & 39.2 & 63.3 & 17.0 & 10.5 & 25.2 & 20.7 & 29.8 & 24.3 & 45.2 & 59.2 \\
     \model & \textbf{40.7} & \textbf{39.6} & \textbf{43.4} & \textbf{67.6} & \textbf{17.6} & \textbf{13.2} & \textbf{30.7} & \textbf{31.4} & \textbf{31.5} & \textbf{33.5} & \textbf{49.5} & \textbf{69.5} \\
    \bottomrule
  \end{tabular}
}
\label{tab:zeroshot_video_retrieval}
\end{table}

\textbf{Video Retrieval.}
We evaluate \model on the video retrieval task. Given a set of videos and related natural language captions, this task requires retrieving the matched video or caption corresponding to its inter-modality counterpart from candidates. We follow the common paradigm to capture visual and text semantics by a visual encoder $f_v(\cdot)$ and a text encoder $f_t(\cdot)$, then calculate the cross-modality similarity matrices as the retrieval guidance. We leverage the multimodal video encoder as $f_v(\cdot)$ and $f_t(\cdot)$ with pretrained ViT-L/14~\cite{dosovitskiy2020image} as the basic CLIP~\cite{CLIP} architecture and finetune the entire model on each retrieval dataset. The training recipes and most of the hyperparameter settings follow CLIP4Clip~\cite{clip4clip}, including training schedule, learning rate, batch size, video frames, maximum text length, etc. To boost model performance, we also adopt the dual softmax loss~\cite{cheng2021improving} as the post-processing operation.

Our model is evaluated on six public benchmarks: MSR-VTT~\cite{xu2016msr}, MSVD~\cite{wu2017deep}, LSMDC~\cite{anne2017localizing}, DiDeMo~\cite{rohrbach2015dataset}, ActivityNet~\cite{caba2015activitynet}, and VATEX~\cite{wang2019vatex}, where we report the results on the standard split following previous works. We measure the retrieval results under the rank-1 (R@1) metric both on text-to-video and video-to-text tasks, which are shown in Table~\ref{tab:video_retrieval}. Results show that our model significantly outperforms all previous methods by a large margin, showing the superiority of \model~ on video-language related tasks. More detailed retrieval results, including rank-5 (R@5) and rank-10 (R@10), can be found in the supplementary materials.

\textbf{Video Question Answering.}
To further demonstrate the vision-language capability of \model~, we evaluate \model~ on video question answering (VQA). Given a video and question pair, VQA is to predict the answer to the question. Unlike the vanilla CLIP model without cross-modality fusion, our multimodal video encoder is able to capture the interaction between modalities with the proposed caption decoder. There are three potential ways to generate features required by the VQA classifier: concatenating the features of the video encoder and text encoder, utilizing the features of the caption decoder only, and concatenating all features from the video encoder, text encoder, and caption decoder. After comparison, we choose to use all three sources of features to boost the performance. The VQA classifier is a three-layer MLP.

We evaluate on three popular public benchmarks: MSR-VTT \cite{xu2016msr}, MSVD \cite{chen2011collecting}, and TGIF \cite{li2016tgif}. We mainly follow the practice in \cite{all_in_one}. The results are shown in Table \ref{tab:vqa} and our model outperforms all previous SOTA, which demonstrates the effectiveness of our cross-modality learner.

\textbf{Visual Language Navigation.} 
Visual-Language Navigation~\cite{anderson2018vision} requires an agent to navigate in unknown photo-realistic environments based on its visual perceptions following natural language instructions. Navigation agents should be capable of capturing spatiotemporal information such as the relative motion of objects from navigation histories, especially when the agent navigates with short step size in continuous spaces. To verify the effectiveness of such ability of our model, we conduct our experiments on the VLN-CE benchmark~\cite{krantz2020beyond}, demanding the agent to function in a continuous environment.

We conduct our experiments using the method proposed in~\cite{an20221st}(CWP-HEP). The history-enhanced planner is a customized variant of HAMT~\cite{chen2021history} which uses a concatenation of depth embedding and RGB embedding as the input embedding. Note that we don't use the tryout controller here since VLN-CE setting allows sliding. This is a strong baseline that already outperforms the previous state-of-the-art method CWP-VLNBERT~\cite{hong2022bridging}. In each decision loop, we collect the latest 16-frame observations to form panoramic navigation videos and then encode the video using ViT-L in \model. The video embedding is concatenated with the RGB embedding and depth embedding as the final image embedding. For evaluation, we refer to~\cite{an20221st} for detailed metrics. \model~could improved our baseline from 50.2\% to 52.9\% in Success Rate(SR) (Table \ref{tab:vlnce}).

\subsubsection{Video Open Understanding Tasks}
\label{sec_open}
\textbf{Zero-shot Action Recognition.}
Zero-shot recognition is one of the extraordinary capabilities of the original CLIP model. With our designed multimodal video encoder, we can also achieve remarkable zero-shot action recognition performance without further optimization. We evaluate our model on Kinetics-400 dataset with 64.25\% accuracy, which outperforms the previous SOTA 56.4\% \cite{Wang2021ActionCLIPAN} by a large margin.
\textbf{Zero-shot Video Retrieval.}
We compare \model~ with CLIP on zero-shot text-to-video and video-to-text retrieval. For fair comparisons, we use the ViT-L/14 model with the pretrained weights \footnote{https://github.com/openai/CLIP.} of CLIP. Wise-finetuning \cite{wortsman2022robust} and model ensemble are employed to further boost the model performance on zero-shot video retrieval. We empirically find that the optimal number of video frames for zero-shot retrieval is between 4 and 8, and the best-performed frame on each benchmark dataset is yielded via grid search. As illustrated in Table \ref{tab:zeroshot_video_retrieval}, \model~ demonstrates superior retrieval ability across all six benchmark datasets. Besides, Florence \cite{florence} used 900M image-text pair for pretraining and it achieved 37.6 R@1 text-to-video retrieval accuracy on MSR-VTT. In comparison, our model outperforms Florence by 4.1\% with much less training data (14.35M video + 100M image v.s. 900M image). These results reveal the effectiveness of our method in learning the joint video-text feature space during pretraining.

\begin{table}[t]
    \centering
    \caption{{Zero-shot multiple-choice on MSR-VTT and LSMDC.} The gray color indicates those methods with supervised training.}
    \begin{tabular}[t]{l l | l l }
        \toprule
        \multicolumn{2}{c}{MSR-VTT} & \multicolumn{2}{c}{LSMDC} \\
        Method  & Accuracy & Method & Accuracy\\
        \midrule
        \color{gray}{JSFusion}~\cite{yu2018joint}  & \color{gray}{83.4} & \color{gray}{JSFusion}~\cite{yu2018joint} & \color{gray}{73.5}\\
        \color{gray}{ActBERT}~\cite{actbert} & \color{gray}{85.7} & \color{gray}{MERLOT}~\cite{merlot} & \color{gray}{81.7}\\
        \color{gray}{ClipBERT}~\cite{clipbert} & \color{gray}{88.2} & \color{gray}{VIOLET}~\cite{violet} & \color{gray}{82.9}\\
        All-in-one~\cite{all_in_one} & 80.3 & All-in-one~\cite{all_in_one} & 56.3 \\
        \model~ & \textbf{93.4}$_{(+13.1)}$ & \model~ & \textbf{77.3}$_{(+21.0)}$ \\
        \bottomrule
    \end{tabular}
    \label{tab:zs_mc}
\end{table}
\begin{table}[t]
\centering
\caption{Results of open set action recognition on two different open sets where the samples of unknown class are from HMDB-51 and MiT-v2, respectively. We report Open Set AUC at the threshold determined by ensuring 95\% training videos (UCF101) are recognized as known.}
\begin{tabular}{clll}
    \toprule
    \multirow{2}{*}{Method} & \multicolumn{2}{c}{Open Set AUC (\%)} & \multirow{2}{*}{Closed Set Accuracy (\%)} \\
                            & UCF-101 + HMDB-51         & UCF-101 + MiT-v2         &                    \\
    \midrule
    OpenMax \cite{bendale2016towards}                    & 78.76         & 80.62         & 62.09      \\
    MC Dropout \cite{gal2016dropout}                & 75.41         & 78.49         & 96.75      \\
    BNN SVI \cite{krishnan2018bar}                    & 74.78         & 77.39         & 96.43      \\
    SoftMax                    & 79.16         & 82.88         & 96.70      \\
    RPL \cite{chen2020learning}                       & 74.23         & 77.42         & 96.93      \\
    DEAR \cite{bao2021evidential}                      & 82.94         & 86.99         & 96.48      \\
    \model  & \textbf{85.48}$_{(+2.54)}$         & \textbf{91.85}$_{(+4.86)}$         & \textbf{97.89}$_{(+1.41)}$          \\
    \bottomrule
\end{tabular}
\label{osar}
\end{table}

\textbf{Zero-shot Multiple Choice.}
Zero-shot multiple choice is another zero-shot task that can demonstrate the model's generality. Multiple choice task aims to find the correct answer in the given choices, usually a small subset such as 5 words. We find that co-training with image-text pairs, wise-finetuning, and the ensemble is essential for the performance on zero-shot multiple choice. We report the zero-shot multiple-choice results in Table \ref{tab:zs_mc} on MSR-VTT and LSMDC datasets. We use the zero-shot performance as a handy indicator for generality in training, and the results show that our model is robust and effective.

\textbf{Open-set Action Recognition.}
In open-set action recognition (OSAR), the model is required to recognize the known action samples from training classes and reject the unknown samples that are out of training classes. 
Compared with images, video actions are more challenging to be recognized in an OSR setting than images due to the uncertain temporal dynamics, and static bias of human actions \cite{bao2021evidential}. Our \model~generalizes well to unknown classes that are out of training classes and outperforms the existing method \cite{bao2021evidential} without any model calibration.

We use the ViT-H/16 model of \model~as a backbone, and finetune it with a simple linear classification head on UCF-101 \cite{soomro2012dataset} training set. To enable \model~to ``know unknown", we follow the method DEAR proposed in \cite{bao2021evidential} and formulate it as an uncertainty estimation problem by leveraging evidential deep learning (EDL), which provides a way to jointly formulate the multiclass classification and uncertainty modeling. Specifically, given a video as input, the Evidential Neural Network (ENN) head on top of a \model~backbone predicts the class-wise evidence, which formulates a Dirichlet distribution so that the multi-class probabilities and predictive uncertainty of the input can be determined. During the open-set inference, high-uncertainty videos can be regarded as unknown actions, while low-uncertainty videos are classified by the learned categorical probabilities. 

\model~can not only recognize known action classes accurately but also identify the unknown. Table \ref{osar} reports the results of both closed-set (Closed Set Accuracy) and open-set (Open Set AUC) performance of \model~and other baselines. It shows that our \model~consistently and significantly outperforms other baselines on both two open-set datasets, where unknown samples are from HMDB-51~\cite{hmdb51} and MiT-v2 \cite{monfort2021multi}, respectively.

\section{Concluding Remarks}

In this paper, we propose a versatile and training-efficient video foundation model \model. To our best knowledge, \model~is the first work to perform best among existing researches on all action understanding, video-language alignment, and video open understanding tasks. Compared with previous related work \cite{maskfeat,mtv,merlot}, it greatly lifts the generality of video foundation models to a new level, by achieving state-of-the-art performance on nearly 40 datasets covering 10 different tasks. 
The model exploits a unified video representation based on the cross-model learning between masked video learning (VideoMAE) and video-language contrastive modeling along with supervised training. Compared with previous foundation models, it is efficient in training. With simple ViT and its corresponding variants, we achieve generalized video representations with 64.5K GPU hours (A100-80G), while CoCa \cite{coca} requires 245.76K TPU hours (v4). We validate such generalized spatiotemporal representation on a spectrum of applications. With simple task heads (even linear ones) and proper downstream adaption tuning, our video representation demonstrates record-breaking results in all used datasets. Even for zero-shot and open-set settings, our model spectrum still gives consistent and non-trivial performance increases, further proving its generalization and adaption.

\subsection{Limitations} 
Our study shows the effectiveness and feasibility of video foundation models instead of giving brand-new formulations or model designs. It focuses on the current popular video perception tasks and handles videos using clips. Its devise can hardly process long-term video tasks, as well as high-order ones, \eg~anticipating plots from the seen parts of a movie. Gaining the capacity to address these tasks is crucial to further push the generality of video representation learning.

\subsection{Future Work}
To further extend the generality of the video foundation models, we suppose embracing model coordination and cognition is necessary for its studies. Specifically, how to systematically coordinate foundation models trained from different modalities, pretraining tasks, and even varied architectures for a better representation remains open and challenging. There are multiple technical routes to address it, \eg\ model distillation, unifying different pretraining objectives, and feature alignment, to name a few. By exploiting previously learned knowledge, we can accelerate video foundation model development sustainably. 

In the long run, foundation models are expected to master cognitive capabilities more than perceivable ones. Considering its feasibility, we suppose one of its research trends is to achieve large-scale spatiotemporal analysis (long-term \& big scene) from the foundational dynamic perception in the open world, leading to essential cognitive understanding.
Moreover, it has raised a tide that combining foundation models with decision-making to form intelligent agents to explore new tasks. In this interaction, data collection and model training are also automated. The whole process enters a closed loop as the interactive results will adjust agent strategies and behaviors. Our initial experiments (Section \ref{sec_vlt}) on vision-language navigation demonstrate the promising future of integrating video foundation models into Embodied AI.

\section{Broader Impact}

We give a video foundation model spectrum \model. It is able to deliver the state-of-the-art performance on around 40 datasets, capable of action discrimination, video-language alignment, and open understanding. Besides of public data, we also exploit self-collected data from the Internet. The employed queries for gathering data are checked for ethic and legal issues and so are the curated data. The power consumption of training \model~is much lower than CoCa \cite{coca}, only taking up 23.19 \% of CoCa. For further impact studies, we need to explore the bias, risks, fairness, equality, and many more social topics.

{\small
\bibliographystyle{ieee_fullname}
\bibliography{egbib}
}

\end{document}